\documentclass[runningheads]{llncs}
\usepackage{eccv}
\usepackage{eccvabbrv}
\usepackage{graphicx}
\usepackage{booktabs}
\usepackage{tabularx}

\usepackage[accsupp]{axessibility}  
\usepackage[breaklinks,colorlinks,citecolor=eccvblue]{hyperref}   

\usepackage{orcidlink}
\usepackage{multirow}
\usepackage{makecell}

\usepackage{colortbl}

\usepackage{xcolor}
\usepackage{colortbl}
\definecolor{codegreen}{rgb}{0,0.6,0}
\definecolor{codegray}{rgb}{0.5,0.5,0.5}
\definecolor{codepurple}{rgb}{0.58,0,0.82}
\definecolor{backcolour}{rgb}{0.95,0.95,0.92}
\definecolor{darkgreen}{rgb}{0,0.4,0}
\definecolor{cerise}{rgb}{0.871, 0.192, 0.388}
\definecolor{carmine}{rgb}{0.59, 0.0, 0.09}
\definecolor{olive}{rgb}{0.332, 0.418, 0.184}
\definecolor{navyblue}{rgb}{0.496, 0.810, 0.837}
\definecolor{airforceblue}{rgb}{0.36, 0.54, 0.66}

\definecolor{navyblue}{RGB}{30,130,255}
\definecolor{citecolor}{RGB}{30,130,255}
\definecolor{gptgreen}{RGB}{78,173,91}
\definecolor{lightgray}{gray}{0.9}
\definecolor{blanchedalmond}{rgb}{1.0, 0.92, 0.8}

\newcommand{\custompara}[1]{{\vspace{1mm}\noindent\textbf{#1}\xspace}}
\newcommand{\seen}{\textsc{seen}\xspace}
\newcommand{\unseen}{\textsc{unseen}\xspace}
\newcommand{\entity}{\textsc{entity}\xspace}
\newcommand{\query}{\textsc{query}\xspace}

\newcommand{\MODEL}{\textsc{AutoVER}\xspace}

\newcommand{\datasetname}{\textsc{Oven}-Wiki\xspace}

\newcommand{\pz}{\hphantom{0}}

\usepackage{paralist}
\usepackage{mathtools, nccmath}

\begin{document}

\title{Grounding Language Models for Visual Entity Recognition} 

\titlerunning{Grounding Language Models for Visual Entity Recognition}

\author{Zilin Xiao\inst{1} \and
Ming Gong\inst{2} \and
Paola Cascante-Bonilla\inst{1} \and \\
Xingyao Zhang\inst{2} \and
Jie Wu\inst{2} \and
Vicente Ordonez \inst{1}
}

\authorrunning{Z. Xiao et al.}

\institute{Department of Computer Science, Rice University, USA \\ \email{\{zilin, pc51, vicenteor\}@rice.edu} \and
    Microsoft STCA, China \\
    \email{\{migon, xingyaozhang, jiewu1\}@microsoft.com}
}

\maketitle

\begin{abstract}

We introduce \MODEL, an Autoregressive model for Visual Entity Recognition. 
Our model extends an autoregressive Multimodal Large Language Model by employing retrieval augmented constrained generation. 
It mitigates low performance on out-of-domain entities while excelling in queries that require visual reasoning. 
Our method learns to distinguish similar entities within a vast label space by contrastively training on hard negative pairs in parallel with a sequence-to-sequence objective without an external retriever.
During inference, a list of retrieved candidate answers explicitly guides language generation by removing invalid decoding paths.
The proposed method achieves significant improvements across different dataset splits in the recently proposed \datasetname benchmark with accuracy on the \textsc{Entity} \seen split rising from 32.7\% to 61.5\%. 
It demonstrates superior performance on the \unseen and \textsc{query} splits by a substantial double-digit margin, while also preserving the ability to effectively transfer to other generic visual question answering benchmarks without further training.


\end{abstract}

\section{Introduction}
 Multimodal Large Language Models (MLLMs) have demonstrated superior performance on a variety of vision-and-language tasks such as visual question answering, image captioning, zero-shot image classification, among others~\cite{peng2023kosmos2, DBLP:conf/icml/0008LSH23}. 
Their abilities can be transferred with few-shot tuning~\cite{tsimpoukelli2021multimodal, alayrac2022flamingo} or even learning in context without parameter updates~\cite{bar2022visual, Wang_2023_CVPR}. Given their remarkable generalization abilities and prior knowledge based on large-scale pre-training, we consider there is still great potential for using them on tasks that require knowledge grounding. 

A recently proposed task requiring knowledge grounding is the Open-domain Visual Entity Recognition (\datasetname)~\cite{hu2023opendomain} task. In this task, given an input image and a question about the image, the goal is to answer with a very specific entity from Wikipedia.
For instance, the answers could be a specific model of airplane, \eg~\textsc{ATR 42}, or \textsc{British Aerospace 146}. 
This is a very challenging task where MLLM-based solutions are prone to {\em hallucinations}, or producing answers that are not at the right level of granularity.  
Importantly, visual entity recognition requires recognizing entities that never appear in the training data (\textsc{unseen} split). 
Moreover, a portion of the \datasetname benchmark (\textsc{query} split) extends the task beyond recognition, requiring non-trivial reasoning to resolve the query question. 

\begin{figure*}[t]
    \centering
    \includegraphics[width=\textwidth]{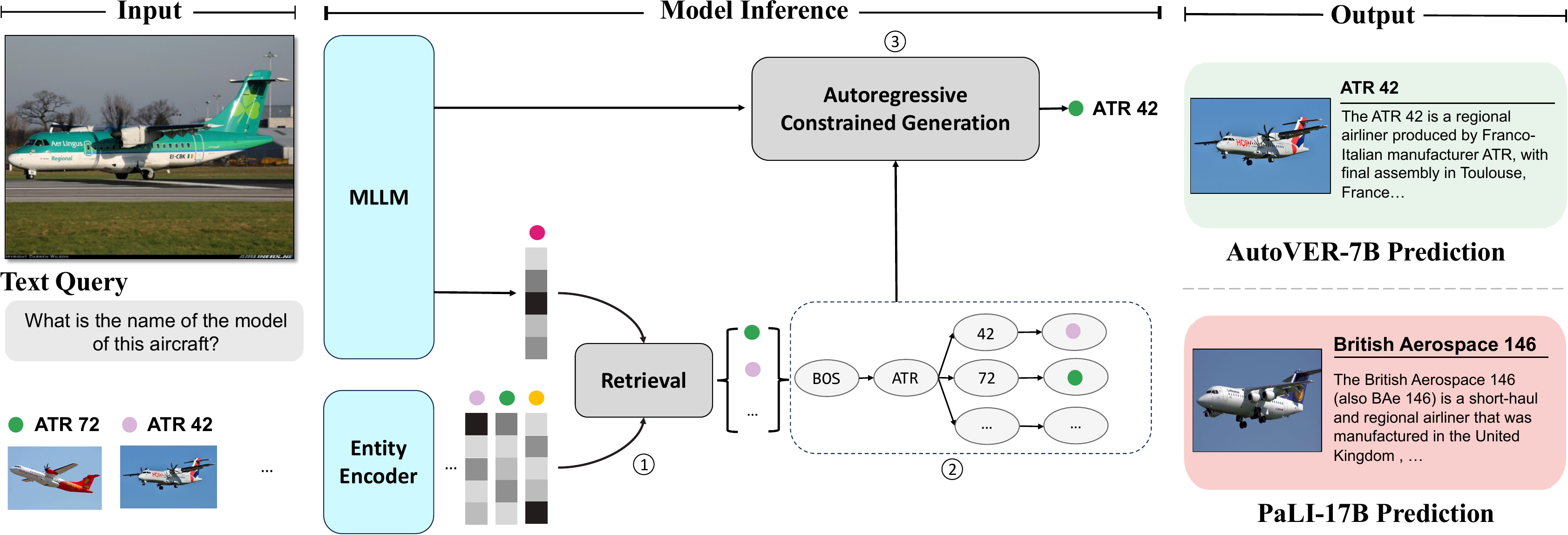} \\
    \caption{A representative query-entity pair from \datasetname. We briefly illustrate the model inference process and compare predictions from \mbox{PaLI-17B} (in red) and \mbox{\MODEL-7B} (in green) which obtains the correct answer: {\sc ATR~42}. 
    \MODEL retrieves entity candidates without an external retriever (step 1), dynamically constructs a prefix tree (trie) (step 2), and performs decoding-time augmentation to guide autoregressive generation (step 3).
    }
    \label{fig:dataset_and_model_overview}
    \vspace{-2mm}
\end{figure*}

Visual~Entity~Recognition~(VER) presents unique challenges compared to general Visual~Question~Answering (VQA)~\cite{malinowski2014multi,agrawal2015vqa}. 
The first challenge lies in the answer label space comprising over 6 million Wikipedia entities, which is prohibitive for classifier-based VQA models~\cite{NEURIPS2018_67d96d45} as many entities considered in this ontology are visually very similar. 
The second issue involves generation-based VQA solutions~\cite{DBLP:conf/icml/0008LSH23, gao2022transform}, where {\em hallucinations} can lead to generated text that is not grounded to the entity space. 
Lastly, existing VQA approaches fail to consider the visual information of candidates~\cite{Shao_2023_CVPR, NEURIPS2023_47393e85}. Entity images from the knowledge base also play a significant role when disambiguating between entities with similar identifiers but different visual appearance. 
These methods also struggle with out-of-domain generalization and multi-hop reasoning, challenges which are covered by the \textsc{unseen} and \textsc{query} split of the \datasetname dataset.

In this work, we introduce an \underline{Auto}regressive \underline{V}isual \underline{E}ntity \underline{R}ecognizer (AutoVER), the first approach that enables multimodal language models to perform accurate visual entity recognition over a massive knowledge base.
\MODEL addresses entity recognition by reformulating the problem as a sequence-to-sequence generation problem, as depicted in~\cref{fig:dataset_and_model_overview}. 
The query image is translated into the token embedding space using a learnable projection layer, in a manner akin to treating images as a foreign language~\cite{liu2023llava, wang2023image}. 
This allows us to utilize pre-trained multi-modal language models and enormously improves performance on the \textsc{query} split of \datasetname which requires reasoning over spatial relations, commonsense, and other visually-situated contexts.

To take advantage of visual clues on the entity side and enhance the generalization capability of the model on \textsc{unseen} entities, we propose a unified and compact retrieval-augmented generation (RAG) framework upon \MODEL.
Specifically, a special token \texttt{<ret>} is added to the vocabulary, whose last-layer hidden states serve as the representation for the query side. A lightweight two-layer Transformer is responsible for fusing visual features from entity images and textual features from entity descriptions and producing a representation for the entities, allowing contrastive learning with query-entity positive pairs in parallel with language modeling. 
Unlike other RAG systems that directly use retrieved items as context~\cite{ram2023incontext, wang2023learning} or infuse them into the model's intermediate hidden states~\cite{NEURIPS2020_6b493230, pmlr-v119-guu20a, fevry-etal-2020-entities}, 
\MODEL dynamically constructs a prefix tree (trie) from the retrieved entity identifiers, and then generates entity identifiers by leveraging a methodology that constrains the next possible tokens and eliminates invalid options based on the trie, thus ensuring the generated text can always be grounded in retrieved candidates.
To alleviate the entity granularity problem, two hard negative sampling strategies are proposed in our contrastive-generative framework to encourage the model to maximally distinguish between similar entities.

In summary, \MODEL offers several advantages over baselines and prior work on VQA in the realm of visual entity recognition: 
(i) It refines the recognition process by integrating contrastive training into an MLLM. 
(ii) The proposed retrieval-augmented constrained decoding framework guarantees correct grounded entity prediction and enhances prediction over \textsc{unseen} entities;
(iii) \MODEL leverages pre-trained visual models and knowledge graphs for hard negative mining which significantly strengthens our contrastive-generative framework in fine-grained entity recognition;
(iv) Experimental results show that \MODEL consistently outperforms fine-tuned CLIP and PaLI variants on all \datasetname splits. 
For instance, \MODEL-7B achieves a 61.5\% accuracy on the \datasetname \entity \textsc{seen} split over PaLI-17B's 30.6\%, and 21.7\% on \entity \textsc{unseen} over PaLI-17B's 12.4\%. 
Additionally, we evaluate \MODEL alongside public multimodal LLMs on entity-related questions from the A-OKVQA dataset, demonstrating that our model can effectively zero-shot transfer to out-of-domain VQA datasets beyond \datasetname.
Furthermore, ablation studies confirm the effectiveness of the introduced retrieve-generate framework.
Code is released at \url{https://github.com/MrZilinXiao/AutoVER}.

\section{Related Work}

\custompara{Visual Entity Recognition} (VER)~\cite{hu2023opendomain} is an emerging task for assessing the ability of a model to perform multi-modal knowledge grounding~\cite{sigir20, DBLP:conf/iclr/ZhangHS22, ouyang2023ontology}.
It can be regarded as a variant of visual question answering~\cite{malinowski2014multi,ren2015exploring,agrawal2015vqa} with the key distinction that the choice set will comprise all entities in a specific knowledge base. 
Earlier solutions view this problem as an image-text-to-image-text retrieval task, where CLIP-based models~\cite{alec2021clip} are fine-tuned and the top-scored answer in inference is treated as the model prediction. 
Alternatively, others rely on fine-tuning generative language models such as PaLI~\cite{chen2022pali} and GiT~\cite{DBLP:journals/tmlr/WangYHLLGLLW22, caron2024generative} in an attempt to match the generated text with candidate entity identifiers using sparse matching approaches such as BM25~\cite{robertson2009probabilistic}. 
Entity Linking (EL) has been a long-standing text-only counterpart to VER, which entails locating mentions in the document and disambiguating them against a set of candidate entities. 
Our method bears the closest resemblance with the recent generative EL paradigm~\cite{DBLP:conf/iclr/CaoI0P21, mrini-etal-2022-detection, xiao2023instructed, xiao2023coherent}, which also reduces the knowledge-grounding problem into a sequence-to-sequence generation task. 
However, we stand out with several distinct advantages, including the seamless integration of retrieval capabilities into the language model and the use of explicit guidance for sequence generation. 

\custompara{Multimodal LLMs} (MLLMs) are motivated by the remarkable reasoning abilities of Large Language Models (LLMs).
LLaVA~\cite{liu2023llava} builds upon LLaMA~\cite{touvron2023llama} and uses instruction-tuning to align visual features to the language space. 
The common practice of prompting MLLMs to produce proxy representations for downstream usage is to expand the vocabulary of language models. Such expansion augments the MLLM functionality beyond language generation. FROMAGe~\cite{koh2023grounding}, GILL~\cite{koh2023generating}, GenLLaVa~\cite{hernandez2024generativevisualinstructiontuning} and LISA~\cite{lai2023lisa} achieve satisfactory results over image-text-to-text retrieval, image generation and reasoning segmentation by inserting special tokens to augment the MLLMs original functionality. 
Our approach, built upon one of the latest MLLM model architecture, presents a compelling alternative to the bi-encoder shallow dot-product interaction or encoder-decoder sparse surface form matching employed in previous works on visual entity recognition. Our method also integratives contrastive learning on the outputs of entities and question pairs, analogous to the image-text contrastive learning used in CLIP-like models~\cite{alec2021clip,shrivastava2023clip,pan2022contrastive} and among visual features used for self-supervised contrastive learning~\cite{chen2020simple,hernandez2023vic,aubret2024self,fan2023contrastive}.

\custompara{Retrieval-Augmented Language Models} (RALMs) represent a class of NLP solutions focusing on knowledge-intensive tasks.
RALMs typically condition a language model on relevant documents from a grounding corpus during generation, thereby enhancing the performance in knowledge-intensive language understanding tasks. 
Lewis \etal~\cite{NEURIPS2020_6b493230} jointly fine-tune a retriever with an encoder-decoder model, enabling the community to explore the RALM paradigm in language understanding. 
Guu \etal~\cite{10.5555/3524938.3525306} train a bi-directional variant and also demonstrate superior performance. 
Apart from retrieved items in context, Févry \etal~\cite{fevry-etal-2020-entities} are the first to integrate retrieved entity supervision by injecting intermediate representations. 
While augmented context helps in mitigating the well-known issue of {\em hallucination}, it does not ensure a faithful prediction with respect to an external knowledge base. 
Our retrieve-generate framework is distinct from existing retrieval-augmented methods. \MODEL reduces the candidate entity set from millions to hundreds for generative language models through retrieval~\cite{iscen2024retrievalenhanced} and dynamically constrains language model generation using a prefix tree. 
This framework coincides with the design philosophy of agents~\cite{gu2022dont} in Interactive NLP~\cite{DBLP:journals/corr/abs-2305-13246}, where an agent interacts with the dynamic environment by performing beam-search on all available options.

\section{Methodology}

We first formulate the Visual Entity Recognition problem in~\cref{sec:problem_definition}. Then we present the overview design of \MODEL in~\cref{sec:model_overview}. We illustrate hard-negative mining with knowledge base source and pre-trained visual model in~\cref{sec:hard_negative_mining} and retrieval-augmented constrained decoding process in~\cref{sec:retrieval_augmented_constrained_decoding}.

\subsection{Problem Definition}
\label{sec:problem_definition}
Visual Entity Recognition can be viewed as a multimodal knowledge grounding task, in which the model is required to process an image-text pair input $ x = ( \mathbf{Q}_{\text{im}}, \mathbf{Q}_{\text{t}} )$ and predict an entity $e$.
$\mathbf{Q}_{\text{t}}$ describes the specific intent that prompts the model to ground some entity $e$ in the image $\mathbf{Q}_{\text{im}}$ to a label space $\mathcal{E}$.
Each entity $e \in \mathcal{E}$ is a member of the knowledge base $\mathcal{K}=\{(e, \mathbf{E}_{\text{im}}, \mathbf{E}_{\text{desc}}) \mid e \in \mathcal{E}\}$ where $\mathbf{E}_{\text{desc}}$ is a text description and $\mathbf{E}_{\text{im}}$ is a set of relevant images of the entity. To circumvent trivial solutions for some query questions, all image-text pairs are annotated so the question cannot be correctly answered without the image. 

\subsection{Model Overview}
\label{sec:model_overview}

\begin{figure*}[t]
	\begin{center}
		\includegraphics[width=0.96\linewidth]{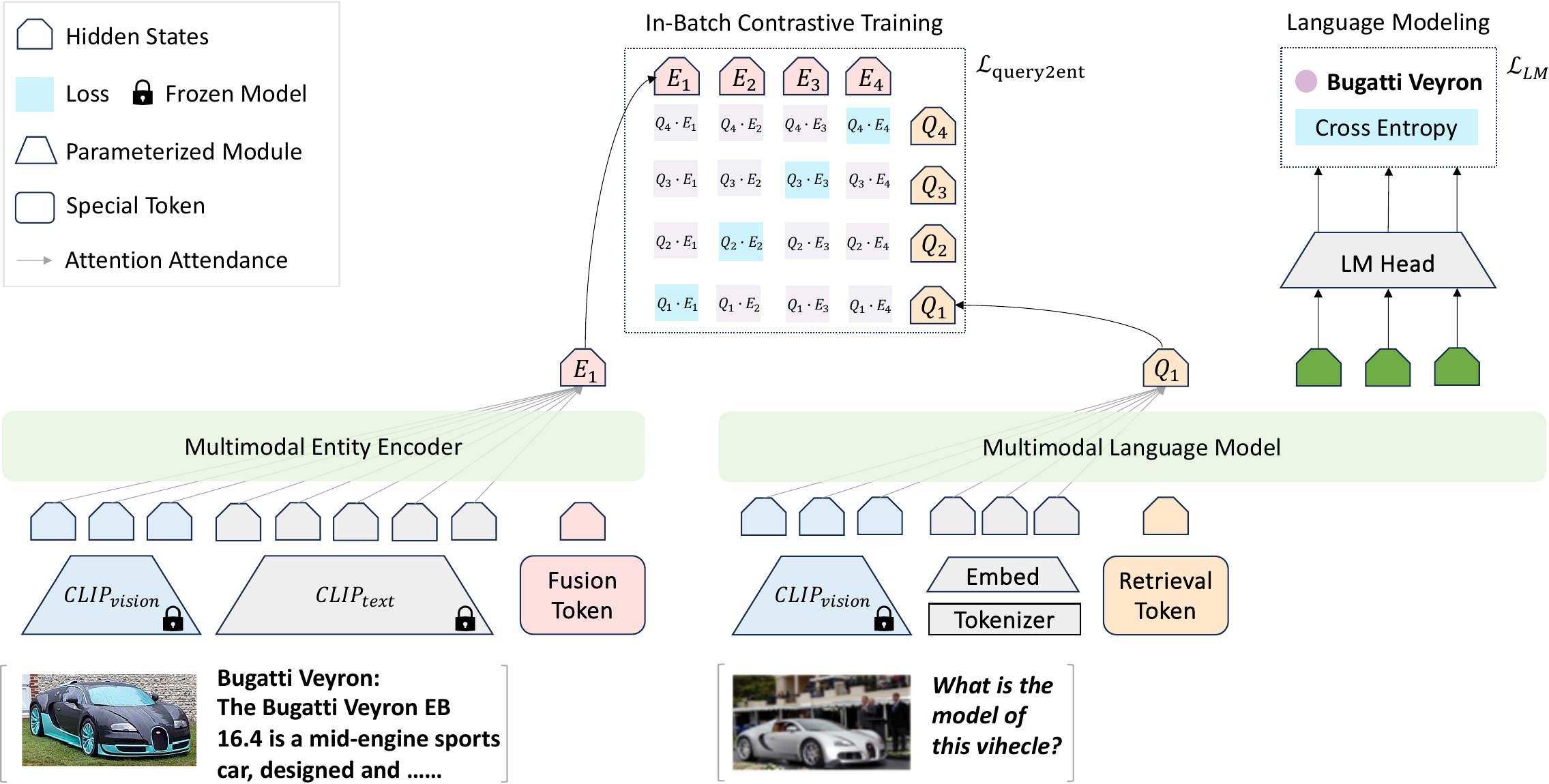}
	\end{center}
	\caption{Joint training of in-batch contrastive learning and language modeling in \MODEL. 
 For each training quadruple consisting of an entity image, an entity description, a query image and a query question, a lightweight Transformer encoder produces the fused entity representation $E_i$ (left half). A special retrieval token prompts the multimodal language model to generate the query representation $Q_i$. The query-to-entity contrastive training ($\mathrm{L}_{\text{query2ent}}$) encourages the correct retrieval of entities given the query pair, and the language modeling ($\mathrm{L}_{\text{LM}}$) helps the successful entity grounding.
 }
	\label{fig:model_train}
\end{figure*}

\MODEL consists of two main modules: a multi-modal language model $f_\phi$ initialized with pre-trained weights $\phi$, and a multi-modal entity encoder $F_\varphi$ that fuses the entity image and textual description. To integrate the retrieval functionality into $f_\phi$, we insert a special token \texttt{<ret>} into the vocabulary of $f_\phi$ with its corresponding token embedding denoted as $\mathbf{H}_\texttt{<ret>}$.

We illustrate the training process in~\cref{fig:model_train}. 
For a training sample $ ( \mathbf{Q}_{\text{im}}, \mathbf{Q}_{\text{t}}, $ $\mathbf{E}_{\text{im}}, \mathbf{E}_{\text{desc}} ) $, we first use a frozen pre-trained CLIP visual encoder $g_\text{vision}$ and a learnable projection $\mathbf{W}_q$ to extract query image features $ \mathbf{H}_{\text{im}}$ and the embedded query text representations $ \mathbf{H}_{\text{t}}$ as follow:
\begin{equation*}
\begin{aligned}
& \mathbf{H}_{\text{im}}=\mathbf{W}_q \cdot g_\text{vision}\left(\mathbf{Q}_{\text{im}}\right), \\
& \mathbf{H}_{\text{t}} = \mathbf{W}_\text{embed} \cdot \mathbf{Q}_{\text{t}},
\end{aligned}
\end{equation*}
where $\mathbf{W}_\text{embed}$ is the token embedding layer associated with $f_\phi$, thus, treating the query image features as a foreign language. We concatenate the embedded query text representations and the query image features with the retrieval token embedding to organize the input instruction to model $ \mathbf{H} = [\mathbf{H}_{\text{im}}, \mathbf{H}_{\text{t}}, \mathbf{H}_\texttt{<ret>}]$.
The last-layer hidden states of \texttt{<ret>} token undergo dimension-matching projection and L2 normalization to a hyperspherical space, serving as the representation on the query side denoted as $Q$. The use of a causal attention mask in $f_\phi$ allows this representation to incorporate the multi-modal query without leaking any label information.

On the entity encoder side, a frozen CLIP visual encoder $g_\text{vision}$ transforms entity image $\mathbf{E}_{\text{im}}$ into grid image features $\mathbf{Z}_{\text{im}}$, and the frozen CLIP text encoder $g_\text{text}$ encodes the entity identifier and description into text features $\mathbf{Z}_{\text{text}}$. 
A two-layer Transformer encoder handles the fusion of two modalities by a fusion token as a soft prompt~\cite{lester-etal-2021-power}, whose design was also adopted in \cite{kirillov2023segment}. The final-layer hidden states of the fusion token are also subjected to dimension matching and L2 normalization, providing the entity representation $E$.

\custompara{In-Batch Contrastive Training.} Given normalized representations from both the query and entity side, $Q$ and $E$, we formally introduce our in-batch contrastive training method specifically designed for query-to-entity retrieval. 
While image-text retrieval has been popular in learning joint vision and language representations~\cite{jia2021scaling, alec2021clip}, few studies have delved into image-text-to-image-text retrieval, which happens to be the focus of query-to-entity retrieval. 
We adopt contrastive learning~\cite{1467314} and the InfoNCE loss~\cite{oord2018representation} used in previous image-text retrieval work. Starting with computing the cosine similarity for a pair of representations,
$$
\operatorname{sim}(Q, E)= \frac{Q \cdot E}{\|Q\| \|E\|},
$$
we minimize the InfoNCE loss for query-to-entity on a mini-batch consisting of $N$ query-entity pairs $(Q_i, E_i)$. Each corresponding pair is considered a positive pair and others are treated as negatives. $\tau$ is a learnable temperature parameter. Our formulation is as follows:
\begin{equation*}
\begin{aligned}
\mathcal{L}_{\mathrm{query} 2 \mathrm{ent}}=-\frac{1}{N} \sum_{i=1}^N\left(\log \frac{\exp \left(\operatorname{sim}\left(Q_i, E_i\right) / \tau\right)}{\sum_{j=1}^N \exp \left(\operatorname{sim}\left(Q_i, E_j\right) / \tau\right)}\right).
\end{aligned}
\end{equation*}
Note that we do not train for the reversed objective, \ie entity-to-query, as it does not align with our retrieval-augmented intuition.

In contrast to image-text retrieval which offers a unique correspondence between images and texts, query-to-entity mapping is subjective, \ie multiple queries in the training data are associated with the same entity. This requires a particular sampling strategy during training to ensure that a batch does not include the same entity, referred to as a ``conflicting batch''. To preserve the integrity of the training data distribution, we implement a rejection sampling~\cite{gilks1992adaptive} method as an alternative to standard random sampling, resorting to resampling upon encountering conflicting batches.

\custompara{Language Modeling. } With organized input instruction $\mathbf{H}$, we optimize the following cross-entropy loss, which is known as the next token prediction loss in causal language modeling:
\begin{equation*}
\mathcal{L}_{\mathrm{LM}}=- \frac{1}{N-n} \sum_{i=n}^N \log P\left(y_i \mid \mathbf{H}, \ldots, y_{i-1}\right), 
\end{equation*}
where $n$ is the length of $\mathbf{H}$, $N$ denotes the length of the concatenation of $\mathbf{H}$ and the expected output, \ie entity text identifier (\textsc{Bugatti Veyron} in Figure~\ref{fig:model_train}). 
Note that we do not backward the next token prediction loss on the input sequence, but only on the target sequence.

The final training loss is a linear combination of the language modeling loss and in-batch contrastive loss, denoted as follows: 

\begin{equation*}
\mathcal{L}=\mathcal{L}_{\mathrm{LM}}+\lambda_r \cdot \mathcal{L}_{\mathrm{query} 2 \mathrm{ent}},
\end{equation*}
where $\lambda_r$ is an empirically determined trade-off hyperparameter.

\begin{figure*}[t]
	\begin{center}
		\includegraphics[width=\linewidth]{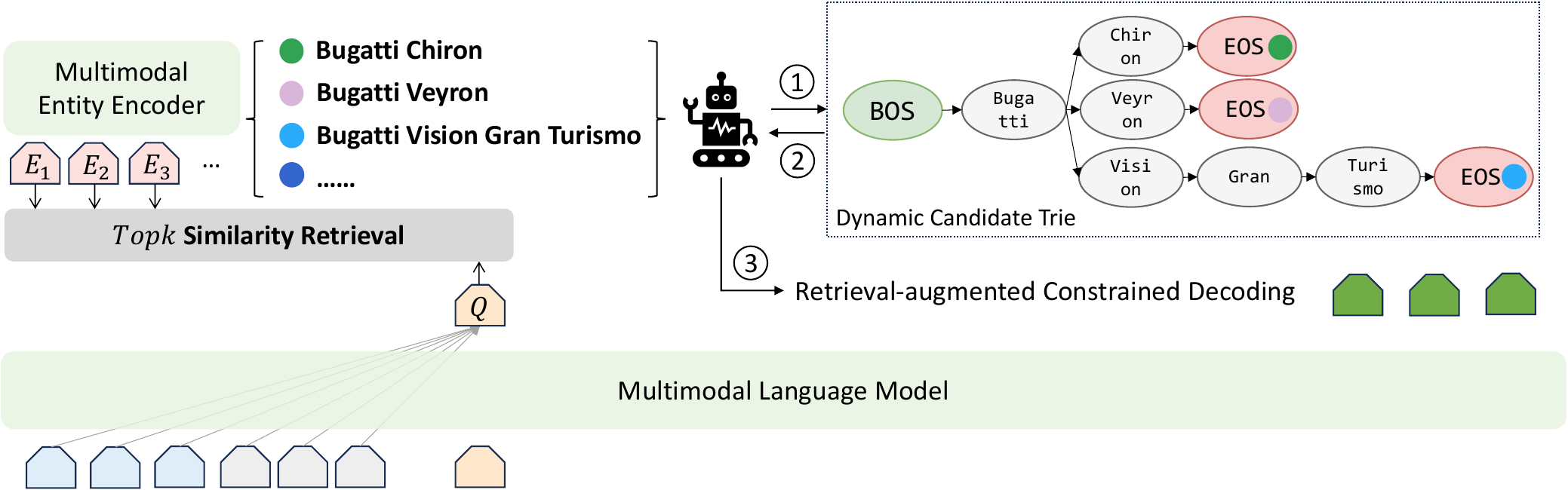}
	\end{center}
 \vspace{-0.1in}
	\caption{Retrieval-augmented constrained decoding illustration of our proposed \MODEL inference process. 
 The representation $Q$ will query a pre-cached entity database constructed using the multimodal entity encoder, and get the top-$k$ candidate entities. 
 A prefix-constrained tree is dynamically built based on retrieved entity identifiers and guides the language model to autoregressively generate the next token, thereby ensuring the successful grounding of generated content.
 }
	\label{fig:model_inference}
 \vspace{-0.1in}
\end{figure*}

\subsection{Hard-Negative Mining}
\label{sec:hard_negative_mining}
To alleviate the entity granularity problem, two hard negative sampling approaches are proposed in our contrastive-generative framework, \ie \textsc{vision-hard} and \textsc{kb-hard}. Both approaches create entity groups used for constructing an in-batch sampler that prefers sampling similar entities in contrastive training, yet they differ in how they generate these similar groups. \textsc{vision-hard} relies on a pre-trained ViT~\cite{DBLP:conf/iclr/DosovitskiyB0WZ21} image classifier to identify visually similar entities based on shared prediction classes. In contrast, \textsc{kb-hard} uses the category hierarchy of Wikidata as an external knowledge source, considering entities that share a parent node in the category hierarchy as knowledge-similar entities. We refer readers to the supplementary material for details about the construction of hard negative groups.

\subsection{Retrieval-augmented Constrained Decoding}
\label{sec:retrieval_augmented_constrained_decoding}
We illustrate the proposed model inference in~\cref{fig:model_inference}. 
After training, all entity candidates are cached using the multi-modal entity encoder $F_\varphi$ to construct an entity vector database $\mathcal{V} \in \mathbb{R}^{n \times d}$ for efficient retrieval. 
Given an evaluation sample $ \left( \mathbf{Q}_{\text{im}}, \mathbf{Q}_{\text{t}} \right) $, we first forward the MLLM for the normalized representation of the \texttt{<ret>} token, and query database $\mathcal{V}$ using top-$k$ similarity search to get $k$ entity candidates. 
The model dynamically generates a prefix tree (trie) that covers $k$ entity candidates, and the trie will explicitly guide entity identifier generation by eliminating impossible decoding paths when autoregressively generating tokens.
The retrieval process takes into consideration the image of entities and improves the model towards out-of-domain entities. Moreover, the retrieval-augmented constrained decoding guarantees the grounding of the generated content to the knowledge base, alleviating the issue of {\em hallucinations}.

\section{Experiments}
We describe the experimental setting in~\cref{sec:experiment_settings} and report the main results in~\cref{sec:main_results}. We present the zero-shot generalization results in~\cref{sec:zero_shot} and ablation study in~\cref{sec:ablation_study} with discussions. For an intuitive demonstration of our model, we present the case study in~\cref{sec:case_study}.

\subsection{Settings}
\label{sec:experiment_settings}

\custompara{Metrics}. 
We follow the standard setup in Hu~\etal~\cite{hu2023opendomain}, which uses accuracy and harmonic mean to evaluate model performance on different data splits. The harmonic mean of each split
will equally weigh the importance of the \seen and \unseen subsets and penalize models that show weakness in either aspect.
Finally, we report the overall harmonic mean on the \textsc{Entity} and \textsc{Query} splits as the final metric for the validation and test sets.

\custompara{Data Pre-processing and Models.} We pre-process all query and entity images by resizing them into 336 $\times$ 336 pixels with padding to keep an identical aspect ratio. 
Entity descriptions are truncated to 77 tokens to fit the context window size of CLIP by cutting off sentences to the maximum available ones. Since query texts are typically short, no truncation is needed for query texts. 
Encoders on the entity side and visual encoder in MLLM are pre-trained \texttt{CLIP-ViTL/14-336px}.
The MLLM is initialized with the \texttt{vicuna-7b-v1.5} or \texttt{vicuna-13b-v1.5} checkpoint and corresponding visual projectors pre-aligned on a 558k subset of LAION~\cite{schuhmann2022laion}, CC~\cite{changpinyo2021conceptual} and SBU~\cite{Ordonez:2011:im2text} curated by Liu~\etal~\cite{liu2023improved}.

\custompara{Training and Evaluation.} We train on the entity train and query train splits of \datasetname, which consists of nearly 5 million query-entity pairs. We conduct all pieces of training in a batch size of 256 on 32 V100-SXM2-32GB GPUs. 
We refer to the supplementary material for hyper-parameter choices and training details. We set $\lambda_r$ to 1 for all training settings.
Due to a limited compute budget, we select 10\% and 50\% of the training data through weighted sampling according to entity occurrence frequency to facilitate ablation studies and reporting final results. For the same reason, we do not run a hyperparameter search. 
In evaluation, the number of retrieved entities $k$ is set to 300 according to empirical trials on the validation split. 

\begin{table*}[t]
	\caption{
	    Comparison among models on the \datasetname \textbf{validation} set. We report accuracies for the \seen and \unseen subsets and the harmonic mean of each split (\textsc{hm}). Metrics of CLIP and PaLI variants are from \cite{hu2023opendomain}. For each subset, \textbf{bold} indicates the best metric.
    }
    \label{tab:main_results_validation}
    \vskip 0.15in
    \centering
    \footnotesize
	\tabcolsep 1.38pt
    {
	    \begin{tabular}{lcl@{\quad}ccc@{\quad}ccc@{\quad}|c}
	      \toprule
	     & & & \multicolumn{3}{c}{Entity Split} & \multicolumn{3}{c}{Query Split} & Overall \\
	      \addlinespace
	      Category && Method & {\seen} & {\unseen} & \textsc{hm} & {\seen} & {\unseen} & \textsc{hm} & \textsc{hm} \\ \midrule
	     \multirow{3}{*}{Discriminative} && $\text{CLIP}_{\xspace\texttt{ViTL14}}$  & \pz5.4 & \pz5.3 & \pz5.4 & \pz0.8 & \pz1.4 & \pz1.0 & \pz1.7 \\
	      && $\text{CLIP Fusion}_{\xspace\texttt{ViTL14}}$ &  32.7 & \pz4.3 & \pz7.7 & 33.4 & \pz 2.2 & \pz 4.2 & \pz5.4 \\
	      && $\text{CLIP2CLIP}_{\xspace\texttt{ViTL14}}$  & 12.6 & 10.1 & 11.2 & \pz4.1 & \pz2.1 & \pz2.8 & \pz4.4 \\
        \midrule

	     \multirow{2}{*}{Generative} && $\text{{PaLI-3B}}$ & 21.6	&6.6 &10.1& 33.2&	14.7&		20.4&		13.5\\
	      && $\text{{PaLI-17B}}$ & 30.6	& 12.4 &		17.6 &  44.2	& 22.4 & 29.8	& 22.1 \\
       \midrule
        \multirow{2}{*}{Zero-shot} && $\text{{BLIP-2}}_{\xspace\texttt{Flan-T5-XXL}}$ & 8.6 & 3.4 & 4.9 & 24.6 & 17.7 & 20.6 & 7.9 \\
            && $\text{{GPT-4V}}$  & 29.8 & 19.3 & 23.4 & 56.5 & {\bf 52.7} & {\bf 54.5} & 32.9 \\

            \midrule
            \multirow{2}{*}{\makecell{Ours}} && $\text{{\MODEL-7B}}$ & 61.5	& 21.7 & 32.1 &  \textbf{69.0}	& 31.4 & 43.2	& 36.8 \\
            && $\text{{\MODEL-13B}}$ & \textbf{63.6} & \textbf{24.5} & \textbf{35.6} & 68.6 & 32.3  & 43.9	& \textbf{39.2} \\
       
	     \bottomrule
	    \end{tabular}
	}
 \vskip -0.1in
\end{table*}

\subsection{Main Results}
\label{sec:main_results}

\definecolor{bggray}{RGB}{230,230,230}

\begin{table*}[t]
	\caption{
	    Results of methods on the \datasetname \textbf{test} set and \textbf{human evaluation} set. Human evaluation results from \cite{hu2023opendomain} are highlighted in \colorbox{bggray}{gray}.
    }
    \label{tab:main_results_test}
    \centering
        	\tabcolsep 3.5pt
    	    \begin{tabular}{l@{\;\;\;}cccccccc}
    	     \toprule
    	     & \multicolumn{2}{@{\;}c@{\;}}{Entity Split} & \multicolumn{2}{@{\;}c@{\;}}{Query Split} & Overall & \multicolumn{3}{c}{Human Eval}\\ 
          \addlinespace
    	  Method & {\seen} & {\unseen} & {\seen} & {\unseen} & \textsc{hm} & {\seen} & {\unseen} & \textsc{hm} \\
        \midrule
                            Human$\,$+$\,$Search \cellcolor{lightgray} & \cellcolor{lightgray} -
    	     & \cellcolor{lightgray} - & \cellcolor{lightgray} -
    	     & \cellcolor{lightgray} - & \cellcolor{lightgray} - & 76.1 \cellcolor{lightgray} & 79.3 \cellcolor{lightgray} & 77.7
    	     \cellcolor{lightgray} \\
          
           \midrule
            $\text{CLIP}_\texttt{ViTL14}$ & \pz5.6 & \pz4.9 & \pz1.3 & \pz2.0 & \pz2.4 & \pz4.6 & \pz6.0 & \pz5.2 \\
            $\text{CLIP Fusion}_\texttt{ViTL14}$  & 33.6 & \pz4.8 & 25.8 & \pz1.4 & \pz4.1 & 18.0 & \pz2.9 & \pz5.0\\
    	   $\text{CLIP2CLIP}_\texttt{ViTL14}$ & 12.6 & 10.5 & \pz3.8 & \pz3.2 & \pz5.3 & 14.0 & 11.1 & 12.4\\  
            \midrule
	      $\text{{PaLI-3B}}$ & 19.1	&6.0& 27.4&	12.0& 11.8& 30.5&	15.8& 20.8 \\
	        $\text{{PaLI-17B}}$ & 28.3 & 11.2 & 36.2 & 21.7 & 20.2 & 40.3 & 26.0 & 31.6 \\
            $\text{{GER-400M~\cite{caron2024generative}}}$ & 31.5 & 17.7 & - & - & - & - & - & - \\
            $\texttt{llava-v1.5-7b}$~\cite{liu2023llava} & 7.5 & 2.1 & 41.9 & \textbf{37.4} & 6.2 & - & - & - \\
            
            \midrule
         $\text{{\MODEL-7B}}$ & 62.8	& 16.0 & 63.7  & 31.9 & 31.8 & 64.7 &	39.9  & 49.4 \\
	      $\text{{\MODEL-13B}}$ & \textbf{65.0} & \textbf{18.6} & \textbf{65.7} & 32.0 & \textbf{34.6} & \textbf{68.4} & \textbf{44.2} & \textbf{53.7} \\

    	     \bottomrule
    \end{tabular}
    \vskip -0.0in
\end{table*}

Results on the validation set are presented in~\cref{tab:main_results_validation}.
Our experimental results demonstrate a consistent improvement in all data splits and subsets on \datasetname. 
Specifically, we observe a double accuracy improvement in the \entity \seen subsets of both \datasetname validation and test splits. 
This indicates that \MODEL can more effectively utilize its parameter capacity, achieving stronger in-domain visual entity recognition capabilities with fewer model parameters. 
The improvement on the \query \seen subsets is slightly lower compared with the \entity \seen improvement but still significant with +24.8\% relative accuracy difference.
We attribute this to the inherent reasoning and visual localization capabilities within the MLLM, a proficiency that has been extensively demonstrated across various benchmarks targeting MLLMs~\cite{naveed2023comprehensive, chen2023large}.

On \unseen subsets, accuracy is severely impacted by queries whose answers involve out-of-domain entities, particularly in the Entity split. Nevertheless, \MODEL still outperforms the largest model, PaLI-17B, which can be attributed to the retrieval-augmented framework eliminating many improbable options for the decision-making of the MLLM. 

We report results on test and human evaluation set in~\cref{tab:main_results_test}. Similarly, \MODEL-13B achieves the best performance on subsets and splits of the test set. 
However, it must be acknowledged that there remains a gap between our results and those of Human + Search in the human validation set, particularly in the human \unseen subset, where the difference reaches 35.1\%.

In addition, we report the zero-shot visual entity recognition abilities of the largest BLIP-2~\cite{DBLP:conf/icml/0008LSH23} checkpoint and GPT-4V\footnote{Due to the limited budget, we only conduct experiments on 10\% of Entity Split$_{(\texttt{Val})}$ and 50\% of Query Split$_{(\texttt{Val})}$. Also, the detail of the image is set to low to minimize prompt token consumption.}~\cite{openai2023gpt4v} on the validation set in~\cref{tab:main_results_validation}, whereas no fine-tuning is performed on the training dataset.
For BLIP-2, zero-shot generated outputs are grounded to Wikipedia using BM25, following generative baseline methods. 
As for GPT-4V, its elaborate generation, a result of instruction tuning, impedes effective evaluation with BM25. 
Hence, we adopt a partial matching evaluation strategy, treating the presence of the entity identifier in the generated text as a true positive.
While BLIP-2 performance is anticipated, as its lack of fine-tuning hinders the acknowledgment of entity grounding intents, GPT-4V has surprisingly exceptional performance on the \query split. 
Nevertheless, GPT-4V poor performance on the \entity split still leaves it trailing behind \MODEL. Furthermore, the opacity of its training data prevents us from determining if GPT-4V excellent zero-shot performance is attributable to possible data leakage.

\subsection{Zero-shot Generalization Results}
\label{sec:zero_shot}
Although \datasetname has already incorporated 14 classic datasets\footnote{ImageNet21k-{P}~\cite{russakovsky2015imagenet,ridnik2021imagenet}, iNaturalist2017~\cite{van2018inaturalist}, Cars196~\cite{krause2013cars196}, SUN397~\cite{xiao2010sun}, Food101~\cite{bossard14food}, Sports100~\cite{sports100}, Aircraft~\cite{maji13fine-grained}, Oxford Flower~\cite{nilsback2008automated}, Google Landmarks v2~\cite{weyand2020google}, VQA v2~\cite{goyal2017making}, Visual7W~\cite{zhu2016visual7w}, Visual Genome~\cite{krishna2017visual}, OK-VQA~\cite{marino2019ok}, Text-VQA~\cite{singh2019towards}} from image recognition and visual question answering, along with extensive annotation efforts that ground answers to the knowledge base, we still hope to test the generalization capability of our method using additional out-of-domain datasets. 
As such, we manually curated a subset from the A-OKVQA~\cite{DBLP:conf/eccv/SchwenkKCMM22} validation dataset. 
The subset does not overlap with any dataset sources in \datasetname and covers all question-answer pairs in the A-OKVQA validation split whose answer can be grounded to entities in the Wikipedia knowledge base. We name the subset \textsc{A-OKVQA-Ent} to emphasize this distinct property. 
Following the design of \datasetname, we divide the subset into \seen and \unseen splits depending on whether the answer entity is in the \datasetname train set.
This subset includes 478 entries out of 1,145 in the A-OKVQA validation set, of which 322 are identified as \seen, while the remaining 156 are classified as \unseen. 

To adhere to the evaluation setting of A-OKVQA, we adopted two evaluation approaches -- \textbf{multi-choice} and \textbf{entity match}. 
In the \textbf{multi-choice} setting, \MODEL constructs a prefix tree with four available options to guide generation, while for other models, options are numbered in the prompt and either the correct option number or an exact entity identifier match in the response is considered a true prediction. 
In the \textbf{entity match} setting, a partial match of the entity text identifier within the generated response qualifies a correct prediction. 
To ensure equitable comparison, baselines included are other generative models comparable in size with \MODEL-7B, specifically LLaVA~\cite{liu2023llava} and its improved \texttt{v1.5} version~\cite{liu2023improved}, OpenFlamingo~\cite{awadalla2023openflamingo}, and InstructBLIP~\cite{dai2023instructblip}. While we expect that all models have not seen the A-OKVQA dataset so that we can assess the zero-shot generalization ability, it is important to acknowledge that LLaVA-v1.5 has been fine-tuned with 50k A-OKVQA multi-choice instructions, potentially giving it an edge. For context, we still include LLaVA-v1.5 results although they are not directly comparable.

\begin{table}[t]
\caption{Model accuracies on A-OKVQA-\textsc{Ent} under \textbf{multi-choice} evaluation. Here methods are provided with multiple choices for answers in the prompt.
($^*$)~Denotes unusual results due to failure to adhere to the provided multiple-choice prompts. 
\textbf{Bold} indicates the best metric under the same setting.
}
\label{tab:aokvqa_mc_results}

\centering
\tabcolsep 3.5pt
\small
\begin{tabular}{clccc}
\toprule
\textbf{Supervision} & \textbf{Method} & \seen & \unseen & Overall \\ 
\midrule
\multirow{4}{*}{\textit{Zero-shot}} & OpenFlamingo-9B$^*$ & 5.9 & 9.0 & 6.9 \\ 
 & $\text{InstructBLIP}_{\xspace\texttt{vicuna-7B}}$ & 53.7 & 49.4 & 52.4 \\ 
 & LLaVA-v1-7B$^*$ & 13.0 & 10.3 & 12.1 \\ 
 & \MODEL-7B (Ours) & \textbf{67.7} & \textbf{52.5} & \textbf{62.8} \\ 
\cmidrule(){1-5}
\textit{Fine-tuned} & LLaVA-v1.5-7B & 72.4 & 73.7 & 72.8 \\ 

\bottomrule
\end{tabular}
\end{table}

\begin{table}[t]
\caption{Model hit rates on A-OKVQA-\textsc{Ent} under \textbf{entity match} evaluation. Here methods are not provided with multiple choices in the prompt and are deemed successful if they produce the right entity anywhere in their output.}
\label{tab:aokvqa_em_results}

\centering
\tabcolsep 3.5pt
\small
\begin{tabular}{clccc}
\toprule
\textbf{Supervision} & \textbf{Method} & \seen & \unseen & Overall \\
\midrule

\multirow{4}{*}{\textit{Zero-shot}} & OpenFlamingo-9B & 39.8 & 30.8 & 36.8 \\ 
& $\text{InstructBLIP}_{\xspace\texttt{vicuna-7B}}$ & 38.8 & 34.0 & 37.2 \\ 
& LLaVA-v1-7B & 45.7 & 36.5 & 42.7 \\ 
& \MODEL-7B (Ours) & \textbf{61.3} & \textbf{42.3} & \textbf{55.0} \\ 
\cmidrule(){1-5}
\textit{Fine-tuned} & LLaVA-v1.5-7B & 47.8 & 42.9 & 46.2 \\ 

\bottomrule
\end{tabular}
\end{table}

We present results on \textsc{A-OKVQA-Ent} in~\cref{tab:aokvqa_mc_results} and \cref{tab:aokvqa_em_results}, and refer readers to the supplementary material for qualitative analysis on this dataset.
Our method still outperforms baseline models across all evaluation settings, with the exception of multiple-choice LLaVA-v1.5 which has been tuned with in-domain instructions. 
Interestingly, we find that LLaVA-v1.5 in the \textbf{entity match} evaluation setting still falls short of \MODEL-7B, although it has seen in-domain samples.
We also observed that LLaVA-v1 and OpenFlamingo, with inadequate instruction tuning, find it challenging to comply with the multiple-choice setting, and instead generate content unrelated to the given options, leading to their significantly lower metrics.
Those findings reveal that generative language models struggle to adapt well to out-of-domain instructions and that unrestricted generation is prone to hallucinations in tasks requiring precise recognition. This further emphasizes the advantages of retrieval-assisted decoding-time augmentation for such tasks.

\subsection{Ablation Study and Discussion}
\label{sec:ablation_study}

We present ablation studies exploring the effect of retrieval augmented generation, constrained decoding, and hard negative mining in~\cref{tab:ablations}. 
Ablations are conducted with 10\% of the training data and as such the metrics in ablations differ from our main results and we refer to our model as \MODEL-7B-0.1. 

\custompara{Retrieval Augmentation.} We first focus on the absence of retrieval augmentation in \MODEL. Without retrieval, the model performs constrained decoding over a prefix tree composed of all Wikipedia entity identifiers instead of a set of retrieved candidates. We notice a slight performance gain on the \seen subset, likely due to the gold entity not being covered in retrieval. 
Conversely, the integrated retrieval design is found to significantly improve the performance of \MODEL on the \unseen subset from non-viable to viable. 
Retrieval-augmented constrained decoding narrows down the decision-making scope of the language model, allowing \MODEL to generate entity identifiers that are never seen during training.

\begin{table}[t]
\caption{Ablation study of \MODEL-7B-0.1 on \datasetname \entity Split$_{(\texttt{Val})}$.}
\centering
\tabcolsep 3.5pt
\begin{tabular}{@{}lccc@{}}
\toprule
Method & \seen  &  \unseen  & \textsc{hm} \\ \midrule
\MODEL-7B-0.1 & 48.9  & 19.0 & 27.4 \\
\quad + w/o retrieval & 50.7  & 0.6	& 1.2 \\
\quad + w/o constrained decoding & 46.8 & 0.6 & 1.2 \\
\quad + w/ LoRA & 43.5 & 2.8 & 5.3 \\ 
\midrule \\[-1em]
\MODEL-7B-0.1-\texttt{[CLS]} & 12.8 & 0.1 & 0.2 \\
\bottomrule
\end{tabular}

\label{tab:ablations}
\end{table}







\custompara{Constrained Decoding.} Upon the excluding of retrieval augmentation, we proceed to disable the entire constrained decoding mechanism, reverting \mbox{\MODEL} to greedily decode the next token, where grounding to an external knowledge base is no longer assured. 
The consequent decrease in \seen accuracy demonstrates that constrained decoding effectively alleviates hallucinations of the model in predicting entities. Unfortunately, since neither checkpoints of encoder-decoder variants from~\cite{hu2023opendomain} in Table~\ref{tab:main_results_validation} and~\ref{tab:main_results_test} nor PaLI pre-trained weights are publicly available, we are unable to assess whether our proposed decoding-time augmenting methods can bring universal improvement on generative baselines. 

\custompara{Parameter-Efficient Tuning (LoRA).} Our ablations extend to the impact of parameter-efficient tuning methods on \MODEL. Specifically, we configure a low-rank adapter (LoRA)~\cite{lowrank} with a rank of 128 and alpha set at 256 and train this LoRA-variant of \MODEL. The results reveal that the efficacy of LoRA falls short of expectation and explicitly harms the recognition performance on the \seen subset. 
We leave more specialized efficient tuning methods for the generative VER framework as future work. 

\custompara{Autoregressive Model or Classifier.} One may attribute the success of \MODEL to the large model size of its underlying MLLM and accompanied expressive capacity instead of our proposed design. In response to that, we devise a \texttt{[CLS]} variant of \MODEL.
It follows the classical design of treating a decoder-only LM as a classifier. Specifically, we introduce a new \texttt{[CLS]} token into the MLLM vocabulary and append this token at the end of each query image-question pair. The corresponding last-layer hidden states are then fine-tuned for classifying the query into the 20,549 existing entities in \datasetname. 
We observed a significant performance degradation in this setting, particularly in the \textsc{unseen} subset. This highlights the ineffectiveness of classifier-based VQA methods in the realm of visual entity recognition, indicating that our model performance stems from more than just its size.

\begin{figure*}[t]
    \centering
    \includegraphics[width=\textwidth]{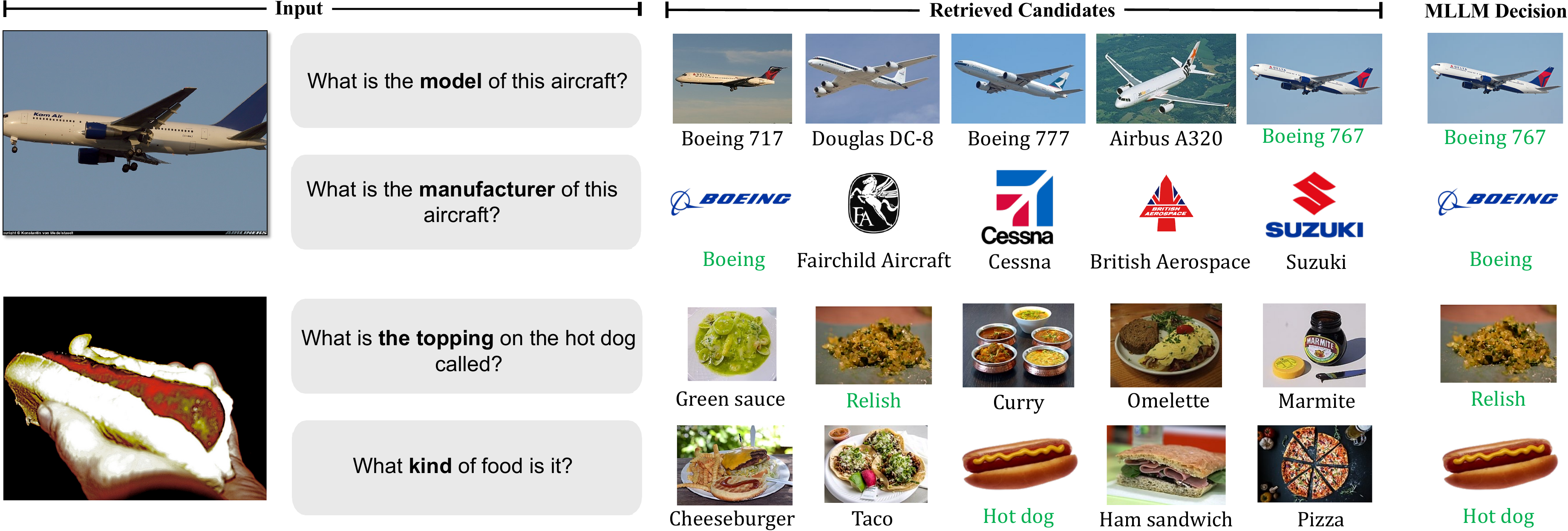}
    \caption{Illustration of selected query image-question pairs, retrieved candidates and \mbox{\MODEL-7B} decisions. \MODEL adeptly captures slight variations in the query text and retrieves entirely different entity candidates, which forms the basis for the generative decisions from the language model.
    }
    \label{fig:long_case_study}
\end{figure*}

\subsection{Case Study} 
\label{sec:case_study}
We illustrate the retrieved entity candidates and model decisions on four representative image-question pairs.
We note that for the same query image, questions with different intents lead to retrieval representations with semantic relevance to the intent, laying a solid foundation for the subsequent constrained generation. 
For instance, in the upper part of~\cref{fig:long_case_study}, the question asking for the specific model of airplane prompts the model to retrieve various kinds of airplanes. With a slight modification from ``model'' to ``manufacturer'' in the query question, the model adapts to retrieve famous airplane manufacturer brands instead of plane models.
Likewise, the model is also proficient in managing queries that demand visual localization and reasoning abilities, as depicted in the lower part of~\cref{fig:long_case_study}. 
We refer to the supplementary material for the error analysis and more case studies with comparisons against GPT-4V. 

\section{Conclusion}
We present \MODEL, a compact retrieval-augmented generation framework specifically designed for visual entity recognition. 
Utilizing a novel constrained decoding technique, this approach effectively overcomes the challenges of low performance in recognizing out-of-domain entities while demonstrating remarkable proficiency in questions that require visually-situated reasoning.
\MODEL marks a significant advancement in visual entity recognition by doubling accuracy on nearly all challenging subsets of benchmarks. We discuss limitations in the supplementary material.

\custompara{Acknowledgements.}
This project was partially funded by an NSF CAREER Award \#2201710, and support from the Ken Kennedy Institute at Rice University. We also thank reviewers for their feedback and encouragement.

\clearpage
\bibliographystyle{splncs04}
\bibliography{custom}

\end{document}